\documentclass{article}

\usepackage{arxiv}

\usepackage[utf8]{inputenc} 
\usepackage[T1]{fontenc}    
\usepackage{hyperref}       
\usepackage{url}            
\usepackage{booktabs}       
\usepackage{amsfonts}       
\usepackage{nicefrac}       
\usepackage{microtype}      
\usepackage{lipsum}
\usepackage{graphicx}
\usepackage[ruled,linesnumbered]{algorithm2e}
\usepackage{amsmath}
\usepackage{url}
\usepackage{setspace}
\graphicspath{ {./images/} }

\title{Algorithms for Optimizing Fleet Staging of Air Ambulances}

\author{
 Joseph Tassone \\
  Department of Computer Science\\
  Lakehead University\\
  Thunder Bay, ON, P7B 5E1 \\
  \texttt{jtasson2@lakeheadu.com} \\
   \And
 Geoffrey Pond \\
  Department of Management\\
  Royal Military College of Canada\\
  Kingston, ON, K7K 7B4 \\
  \And
 Salimur Choudhury \\
  Department of Computer Science\\
  Lakehead University\\
  Thunder Bay, ON, P7B 5E1 \\
}

\begin{document}
\maketitle
\begin{abstract}
In a disaster situation, air ambulance rapid response will often be the determining factor in patient survival. Obstacles intensify this circumstance, with geographical remoteness and limitations in vehicle placement making it an arduous task. Considering these elements, the arrangement of responders is a critical decision of the utmost importance. Utilizing coordinate data of known health and air facilities, this research structured an optimal coverage problem with integer linear programming. For accurate comparison, the Gurobi optimizer was programmed with the developed model and timed for performance. A solution implementing base ranking followed by both local and Tabu search-based algorithms was created. The local search algorithm proved insufficient for maximizing coverage, while the Tabu search achieved near-optimal results. In the latter case, the total vehicle travel distance was minimized and the runtime significantly outperformed the one generated by Gurobi. Furthermore, variations utilizing parallel CUDA processing further decreased the algorithmic runtime. These proved superior as the number of test missions increased, while also maintaining the same minimized distance.
\end{abstract}


\section{Introduction}
\label{sec:introduction}
Rapid disaster response can be the difference in determining a patient's survival. In urban environments, ambulance retrieval is a standard procedure; however, the process becomes increasingly complicated with the remoteness of an incident and the dispersing of a population. As such, placement of responders for optimal area coverage is an important and critical decision. Additionally, many air ambulance services contain a comparatively small fleet to handle vast areas\cite{ornge_2019}. Given such a small contingency, proper placement of these vehicles becomes all the more crucial. 

Multiple solutions have been proposed in previous works, with some implementing near optimal metaheuristics \cite{10.1007/978-3-030-04070-3_17} or concentrating specifically on scheduling \cite{ornge_schedule}. For this research, the problem was formulated to maximize coverage, while minimizing the solution runtime. Known coordinates of air and health facilities aided in developing a more realistic scenario rather than relying on synthetic generation. Each mission began at a potential base, performed a pickup, dropped off a patient, and then returned back to the same base. The primary purpose was determining which bases the vehicles should be placed at to maximize the coverage. While exact methods are an option when time is not a factor, in emergency situations there are instances where vehicles must be repositioned quickly to fill demands. In this case algorithmic metaheuristics are a much more  necessary. Furthermore, many past works only considered sequential implementations, whereas the Compute Unified Device Architecture (CUDA) platform provided an opportunity for further improvement through parallelization. This research aimed to model the problem in terms of integer linear programming and then use custom algorithms to achieve a near-optimal solution.

The remainder of this paper is arranged as follows. Section \ref{sec:related_work} describes the related work, emphasizing previous or similar techniques for resolving the topic. Section \ref{sec:model} presents the problem domain and description, along with the constraints. Section \ref{sec:algorithm} describes the base ranking, local search, and Tabu search-based solutions. Sections \ref{sec:results} and \ref{sec:conclusion} explain the results and conclude the paper.

\section{RELATED WORK}
\label{sec:related_work}
The healthcare industry is only one candidate when considering the optimization of air assets. The present work explores the minimization of total distance for placement; however, others have examined cost in conjunction with distance. Fernandez-Cuesta et al. \cite{doi:10.1080/03155986.2016.1262583} looked at this problem from the perspective of the oil industry and suggested two heuristics for optimizing the position of a fleet of helicopters. Placement of vehicles is considered an NP-hard problem, making it unrealistic to achieve an optimal solution or use purely iterative techniques. Dong et al. \cite{DONG2016195} confronted this issue, taking a relaxed approach by solving for a subset of decision variables, and then locking the solved variables. An elementary approach was then utilized for resolving the remaining variables, minimizing the operation costs while encouraging maximized profits on fleet composition and service levels. 

Regarding fleet management, an optimized solution must provide coverage with a minimized retrieval distance and potentially a minimal resource cost. Using approximate dynamic programming, Schmid \cite{SCHMID2012611} solved a dynamic ambulance reallocation problem. The approach resolved two of the previously mentioned criterions (coverage and cost) by relocating among a fixed set of stations. This had the added benefit of reducing the cost of subsequent ambulance requests. Maleki et al. \cite{MALEKI2014271} attacked a similar relocation problem; however, additionally attempted to minimize the total transit time by ambulances on succeeding calls. In \cite{KOC201633}, the authors confronted the added constraint relating to time-windows. There was a similar objective of providing minimum coverage at a reduced cost, yet the approach instead used a hybrid metaheuristic. Based on mixed-integer programming formulations, a hybrid evolutionary search algorithm (HESA) was developed. The algorithm shared similarities with genetic algorithms, while also using an embedded local search operator for improving offspring generated from the crossover operation.

Empirical data allows a model to make use of past trends for present solutions. Utilizing information like travel time, dispatch delay, and pickup time; McCormack et al. \cite{10.1016/j.ejor.2015.05.040} developed a simulation for land-based ambulances. For actual optimization, the simulation relied on a genetic algorithm for fleet assignment. Similarly, the work of Zhen et al. \cite{ZHEN201412} took a simulation approach with the use of a genetic algorithm for optimization. The work looked to maximize the expected survival probability across variable patient classes. In this approach, the authors utilized the tactic for developing a model for actual deployment and redeployment. As with previously mentioned works, Pond et al. \cite{10.1007/978-3-030-04070-3_17} implemented a genetic algorithm, although directed the solution space towards air ambulance vehicle placement. The paper asserted that population density alone was not an accurate enough determinant for placement of vehicles, and instead relied on a large volume of past data for reaching a resolution. Other generalized solutions for these types of set-covering problems implemented fuzzy parameters, such as those in \cite{HWANG2004861,doi:10.1080/03081079108945020}.

Local and Tabu search are well documented methodologies for resolving optimization and set coverage problems. Literature on these algorithms is substantial and will only briefly be touched upon in this section. In local search, small localized changes are periodically made until a solution is approximately optimal. Tabu search can be seen as an extension of local search-based algorithms. It allows for the exclusion of recently explored areas within a search space and can possibly allow moves that would not improve the objective. A list of previous states are held within a Tabu list, which prevents a search from reaching a local optimum. It only records recent moves and will not allow a solution which has been explored within a particular period. The list clears after a predetermined number of iterations, although the size of the list can vary depending on the problem \cite{EDELKAMP2012633}. In \cite{HALPER2015210}, Zimmermann exploited local search for resolving a mobile facility location problem, whereby clients were assigned to existing facilities so that the total movement and client travel costs were minimized. The model reduced the problem into smaller solvable subproblems and then implemented a modified local search for optimization.  Gendreau et al. designed and modeled an ambulance location problem, resolving it with Tabu search \cite{GENDREAU199775}. The objective was to maximize the coverage using two ambulances, constrained by actual requirements imposed by EMS service laws. Real and randomly generated data points were used, approaching near-optimal results in a reasonable computing time compared to the CPLEX optimizer. Oberscheider and Hirsch looked into efficient transport for non-emergency patients utilizing real-patient data from the Red Cross of Lower Austria \cite{10.1186/s12913-016-1727-5}. They generated all combinations of patient transports, then performed a set partitioning action upon the previous generation to gain an initial solution. They then inputted these combinations into a Tabu Search and optimized the routing.

Implementing parallelization to improve algorithms is not a new trend and has generally been used to speed up calculations through the use of the graphics progressing unit (GPU) \cite{LEE2012175}. Reorganizing algorithms to take advantage of multiple simultaneous threads can dramatically enhance performance and see a huge improvement in the runtime of certain techniques. Hussai et al. altered the particle swarm optimization algorithm through the use of the CUDA platform \cite{7829615}. Through partially coalescing memory accesses, they were able to achieve a massive time improvement when applied to benchmarks. Fabris and Krholing utilized similar benchmarks to test the applications of the CUDA platform on a co-evolutionary differential evolution algorithm for solving min-max optimization problems \cite{FABRIS201210324}. Through this application, they found that the algorithm converged to a near optimal fitness and scaled far better than non-parallel variations. Following review, there is currently little published research on applying GPU parallelization to ambulance problems. Similarly, Schulz et al. discussed in their survey that there is a comparatively small amount of literature on applying GPU parallelization to local or Tabu search \cite{Schulz2013}. Much of the current research has been directed towards swarm algorithms, though the survey suggests that it is still useful for local search-based methods.

As previously discussed in \cite{10.1007/978-3-030-04070-3_17}, multiple maximum coverage problems have relied on population density for the development of an optimized solution space. This is not feasible when considering a sizeable non-uniform density over a large region. Utilizing some traditional methodologies would mean that a significant area is ignored, risking patient survival through invalid placement. The problem can be treated as a coverage problem with the added addition of ensuring equal importance among those in the north. It should be noted that research on the effect of ambulance response time is still a wary topic \cite{10.1080/10903120902935363,10.1136/emermed-2016-206139.29,10.3238/arztebl.2018.0541,PEREZ2015481A}; however, from an economic standpoint, there is an interest in reducing travel distance.

Prior research has relied on genetic algorithms for achieving optimization \cite{10.1007/978-3-030-04070-3_17}, yet was not utilized in this paper as further constraints were implemented and tested upon. Additionally, the methodology of this research was compared against optimized results; achieving near-optimal itself. Similar works have been completed regarding real provided data \cite{ornge_schedule}; although this was more directly related to scheduling, did not list substantial constraints, and made use of a set-partitioning integer program. Given the organization of the data and prior works, this research will explore both local and Tabu search-based solutions, while at the same time assessing the usefulness of parallelizing both algorithms.

\section{MODEL}
\label{sec:model}
Placement of ambulances for maximum coverage is a more nuanced problem that cannot rely solely on demographic data alone. Patients typically move to specialized facilities if the care required is more particular. Additionally, if a region's population is sparse then population density is a poor predictor for developing an optimized solution. As a result, historical mission data can instead be utilized for considering possible future demands. For this research, known medical facility and base coordinates were implemented, consisting of both transfers and area pickups by rotary-wing aircraft.

A mission consists of a pickup, a delivery, and a return to base. To simplify the calculation, the distances between pickup and delivery points are ignored since excluding them does not effect the final coverage determination. As this is not a scheduling problem, the formulation considers that each base can only hold a single aircraft. In general dispersing the vehicles will be more beneficial the more spread out missions become over a large area. In essence this process would be completed prior to scheduling in order to determine the best placement of vehicles for servicing missions. Additionally, cost determination for travel is related primarily to scheduling and less to coverage. The objective function can be modified for cost if the need requires it; however, distance is more useful in determining the best regional coverage. Similarly, vehicle speed is ignored for this problem as some areas are clustered with vehicle specific missions (rotary-wing helicopters are required). In this case speed of the vehicle is irrelevant, as the missions cannot make use of the faster vehicle. 

\medbreak

\noindent The aircraft fleet is made up of following two sets:
\\$R$: the set of all rotary-wing helicopters  $r_i \in R \:\forall\: i = 1, ..., 8$
\\$F$: the set of all fixed-wing planes  $f_j \in F \:\forall\: j = 9, ..., 12$

\medbreak

\noindent The potential bases consist of the following two sets:
\\$A$: the set of all aerodromes capable of supporting both rotary-wing helicopters and fixed-wing planes, with each being a 3-tuple of form

$a_k=<k\:\phi_{k}\:\psi_{k}>,a_k \in A \:\forall \:k= 1, ..., 274$

$k \equiv$ aerodrome ID

$\phi_{k} \equiv$ row coordinate of aerodrome $k$

$\psi_{k} \equiv$ column coordinate of aerodrome $k$
\\$H$: the set of all heliports, only supporting rotary-wing helicopters, with each being a 3-tuple of form

$h_n=<n\:\phi_{n}\:\psi_{n}>,h_n \in H \:\forall \:n = 275, ...,378$

$n \equiv$ heliport ID

$\phi_{n} \equiv$ row coordinate of heliport $n$ location

$\psi_{n} \equiv$ column coordinate of heliport $n$ location

\medbreak

\noindent The set set of all missions consists of:
\\$M$: the set of all missions, with each being a 6-tuple of form

$m_z=<z\:\phi_{p}\:\psi_{p}\:\phi_{d}\:\psi_{d}\:\rho>,m_z \in M \:\forall \:z$

$z \equiv$ mission ID

$\phi_{p} \equiv$ row coordinate of patient pick-up location

$\psi_{p} \equiv$ column coordinate of patient pick-up location

$\phi_{d} \equiv$ row coordinate of patient delivery location

$\psi_{d} \equiv$ column coordinate of patient delivery location

$
\rho =
\begin{cases}
  1 & \text{if mission requires rotary-wing helicopter}\\  
   0 &\text{otherwise}   
\end{cases}
$

\medbreak

\noindent Decision variables consist of the following:
\\$
    v_{im} =
    \begin{cases}
        1 & \text{if rotary-wing helicopter $i$ is assigned to mission $m$}\\
        0 & \text{otherwise}
    \end{cases}
$
\\$
    w_{jm} =
    \begin{cases}
        1 & \text{if fixed-wing plane $j$ is assigned to mission $m$}\\
        0 & \text{otherwise}
    \end{cases}
$
\\$
    x_{jk} =
    \begin{cases}
        1 & \text{if fixed-wing plane $j$ is assigned to aerodrome $k$}\\
        0 & \text{otherwise}
    \end{cases}
$
\\$
    y_{ik} =
    \begin{cases}
        1 & \text{if rotary-wing helicopter $i$ is assigned to aerodrome $k$}\\
        0 & \text{otherwise}
    \end{cases}
$
\\$
    z_{in} =
    \begin{cases}
        1 & \text{if rotary-wing helicopter $i$ is assigned to helipad $n$}\\
        0 & \text{otherwise}
    \end{cases}
$

\medbreak

To perform optimal vehicle placement, distances needed to be calculated. $D$ represented the distance between a potential aerodrome with the sum of each mission's patient pick-up and delivery location. Similarly, $E$ described a measurement applied instead to helipads. For determining optimal distances, there are a number of distance formulas that may be substituted in the model. If the data was purely simulated then Euclidean distance would have been a valid option, though this is not practical when using coordinate data based on latitude and longitude. The Earth is not a perfectly flat space and the curvature must be considered in order to garner an accurate measurement. In this case Haversine distance is a far more reliable metric for the model and can be calculated with the following formula:
\begin{equation}
  \label{e1}
  \resizebox{0.9\columnwidth}{!}{
    $\mathit{D_{ij}} =
    2r\sin^{\:\!\!-1}\left(
    \sqrt{\sin^{\:\!\!2} \left(\frac{ \phi - \phi_{p} }{2}\right) +  \cos(\phi)\cos(\phi_{p})\sin^{\:\!\!2} \left(\frac{\psi - \psi_{p}}{2}\right)}\right)$
  }
\end{equation}

In the above formula $\phi$ represents row location values for $\phi_{k}$ or $\phi_{n}$ (aerodromes and helipads) respectively. Similarly, the same can be said for $\psi$ applying to column location values $\psi_{k}$ or $\psi_{n}$. The remaining distance matrices use the same formula; however, $\phi_{p}$ and $\psi_{p}$ can be substituted for $\phi_{d}$ and $\psi_{d}$. The resulting matrices are both of dimension $z$ (number of missions) by $k$ or $n$ (number of aerodromes or heliports) and referenced for achieving total distances for each mission. The optimization model takes the following form:

\begin{align} 
&\textbf{minimize}\nonumber\\
\label{e2}
&\sum_{m}\Big(\sum_{i}v_{im}\Big(\sum_{k}D_{mk}+\sum_{n}E_{mn}\Big)+\sum_{j}w_{jm}\Big(\sum_{k}D_{mk}\Big)\Big)
\\
&\textbf{subject\: to}\nonumber\\
\label{e3}
&\sum_{i}v_{im} + \sum_{j}w_{jm} = 1 ; \forall \:m\\
\label{e4}
&\sum_{k}y_{ik} + \sum_{n}z_{in} = 1 ; \forall \:i\\
\label{e5}
&\sum_{k}x_{jk} = 1 ; \forall \:j\\
\label{e6}
&\sum_{i}z_{in} \leq 1 ; \forall \:n\\
\label{e7}
&\sum_{j}x_{jk} + \sum_{i}y_{ik} \leq 1 ; \forall \:k\\
\label{e8}
&\sum_{m}v_{im} \leq \sum_{n}z_{in} + \sum_{k}y_{ik} ; \forall \:i\\
\label{e9}
&\sum_{m}w_{jm} \leq \sum_{k}x_{jk} ; \forall \:j\\
\label{e10}
&w_{jm} \leq 1 - \rho_{m} ; \forall \:j\\
\label{e11}
&v_{im}, w_{jm}, x_{jk}, y_{ik}, z_{in} \in \{1,0\}; \forall \:i, j, k, m, n
\end{align} 

The objective function is described by Equation \ref{e2}, where the total distance flown by each aircraft assigned to a mission is minimized. The binary decision variables applied are used with references to matrices $D$ and $E$ respectively. These are used to sum the total distances for each assigned base. The equation is separated into three parts, with each separation indicating the possible assignments that can occur:
\begin{itemize}
  \item Rotary-wing helicopter located at aerodrome assigned to a mission.
  \item Fixed-wing plane located at aerodrome assigned to a mission.
  \item Rotary-wing helicopter located at helipad assigned to a mission.
\end{itemize}
Furthermore, the objective function is constrained based on the rules set by Equations 3 to 11. Each mission can only support a single aircraft, constrained by Equation \ref{e3}. Equations \ref{e4} and \ref{e5} limit each aircraft to a single base, while equations \ref{e6} and \ref{e7} ensure that for every helipad or aerodrome there is at most one vehicle. Enforced by prior constraints, Equation \ref{e8} guarantees that each rotary-wing assigned mission has an occupied base if the variable is set. In this particular case, both decision variables for a base assignment will not be set as a result of previous constraints. Similarly, Equation \ref{e8} does the same type of operation, except towards fixed-wing aircraft and aerodromes. As some missions are rotary-wing only, Equation \ref{e10} prevents a fixed-wing aircraft from being assigned to these missions. Lastly, Equation \ref{e11} fixes the decision variables to a binary format.

\section{ALGORITHM}
\label{sec:algorithm}
Solutions were designed with the previously discussed model while considering the prior constraints and minimization requirements. The description of this solution is described in Algorithm \ref{base_rank}, Algorithm \ref{local_search}, and Algorithm \ref{tabu_search}. All versions had a sequential and parallel CUDA implementation, with minor changes in design. The primary differences related to the lack of outer loops in the CUDA versions, as these indices were acted upon simultaneously by individual threads. Additional differences are given following a description of each algorithm. Given that the differences between the parallel and sequential algorithms are only minor, the actual outlining of them are expressed using the sequential versions.

\subsection{Base Ranking}
Algorithm \ref{base_rank} reduced the problem scope by ranking the most effective bases for covering every mission. The idea was to allow for a more reasonable starting position over the local search, acting on a randomly assigned set. Specifically, the algorithm looked at which bases had the lowest total Haversine distance and assigned a vehicle to each top-ranked base. The number of vehicles was predetermined based on known air ambulance fleet sizes and represented in lines 1 and 2. The input for the solution utilized coordinate data (longitude and latitude) for base locations, mission pickup, and mission destination. In lines 3-5, this information was respectively assigned to the $Destination$, $Pickup$, and $Base$ matrices. In this case, the organization of the matrix indices corresponded to each mission (the exception being $Unused$). As an example: the first index of $Destination$, $Pickup$, and $Base$ would be one complete mission from start to end.

Lines 8-12 of the algorithm summed the Haversine distance for each base relative to both its destination and mission pickup. Following this, lines 13-18 determined the top bases for each mission based on the minimum summed Haversine calculation. For this particular set, 12 bases were chosen as this corresponded to the number of vehicles available for assignment. Per the model's decision variables and Equations \ref{e3}-\ref{e10}, a fixed-wing vehicle could not be assigned to a helipad. As such, lines 14-16 prevented the number of helipads chosen to exceed the number of rotary-wing vehicles. The ranking for each mission was then taken at line 19, where each distance was assessed based on the based on chosen bases in $Top\_Vals$ relative to $Distance$. Each vehicle was then assigned to a respective aerodrome or helipad at lines 20-21. Helipads were allocated first to ensure that fixed-wing aircraft had an available aerodrome available. From lines 22-29 the actual mission to base assignment took place. A constraint handled by the algorithm was that each base selected had to be able to accommodate a specific assigned vehicle. In this case 8 helicopters and 4 planes; meaning that the top bases were occasionally the second-best option instead of the first choice. Additionally, some missions specifically required the use of a rotary-wing aircraft, further constraining the rankings (lines 23-25). Line 24 and 27 finalized the algorithm, assigning the bases to each mission (placing them at the corresponding index in $Base$). This solution did not guarantee the best possible placement, only that the local search had a strong starting point. Swaps among unused bases still needed to be considered, as alternatives may have yielded better results. As such, line 30 assigned any unused bases to $Unused$.

The Base Ranking algorithm operates quite efficiently, although can be easily made parallel by the elimination of the outer loop at line 8. CUDA operates through simultaneous thread organization, so by separating $i$ into individual threads, each can perform the inner loop $j$ at the same time. As there is no race conditions for writing to $Distances$, each can write to a row $i$ without issue. Another modification can be made at line 17, as it requires the summing of every row $i$ in the $Distance$ matrix to determine the total distance of each base to every mission. This would again apply a thread to each row $i$ to determine the respective sums. These modifications remove the bottleneck associated with scaling for an increase in missions. The remainder of the algorithm did not require parallelism, as there were only small quick accesses using single loops.

\medskip
\begin{algorithm}[]
 \KwData{Base, mission pickup, and mission destination data}
 \KwResult{Assignment of vehicles to top aerodromes}
 $heli\_number \leftarrow$\ number of helicopters\;
 $plane\_number \leftarrow$\ number of airplanes\;
 $Destination \leftarrow$\ coordinates of destinations for respective missions\;
 $Pickup \leftarrow$\ coordinates of each mission\;
 $Base \leftarrow$\ coordinates of aerodromes\;
 $Distances \leftarrow$\ empty list sized to length of $Pickup$ by length of $Base$\;
 $Top\_Vals \leftarrow$\ empty list sized to $heli\_number$ + $plane\_number$\;
 \For{$i \gets 1$ to number of bases}   {
    \For{$j \gets 1$ to number of missions}   {
        Add to $Distances$ summed Haversine distance of $Base$ $i$ to each respective mission $Pickup$ and $Destination$ $j$ \;
    }
 }
 \While{$Top\_Vals$ is not full} {
    \If{Number of helipads chosen = $heli\_number$} {
        Choose an aerodrome instead to provide enough locations for $plane\_number$\;
    }
    Populate $Top\_Vals$ with indices of top aerodromes (minimum summed Haversine distance to each mission)\;
 }
 Rank chosen $Top\_Vals$ bases in $Distances$ for each mission\;
 Assign rotary-wing vehicles to helipads\;
 Assign remaining rotary-wing and fixed-wing vehicles to aerodromes\;
 \While{all missions are not assigned a base} {
    \If{mission requires a fixed-wing vehicle} {
        Assign best available and compatible rotary-wing occupied base to the mission\;
    }
    \Else {
        Assign best available and compatible rotary-wing  or fixed-wing occupied base to the mission\;
    }
 }
 $Unused$ = remaining unassigned aerodromes and helipads\;
 \caption{Base Ranking}
 \label{base_rank}
\end{algorithm}
\normalsize

\subsection{Local Search Fleet Optimization}
Algorithm \ref{local_search} accepted the ranked data and unused bases assigned in Algorithm \ref{base_rank}. As there would be no variation on successive runs (and the guarantee of optimal placement), two permutation matrices were generated corresponding to each mission index (line 5 and 6). Essentially, this meant that swaps or take overs would occur at different points each time, offering a distinction between results and allowing the exploration of varying neighbourhoods. Per line 9, the goal of the algorithm was to minimize the total Haversine distance across all missions. The entire set would iterate multiple times completely, stopping only after no further improvement could be found.

The algorithm ran through every mission at least once (line 11), while ensuring that each had a chance to swap with every corresponding option (line 12). Two sets of changes were possible, depending on whether a given unused base contained a vehicle. The choices from lines 13-22 were for the corresponding mission vehicle to be taken over by another vehicle assigned base or moved to a new compatible base (an empty base in $Unused$). This depended again on whether the base in $Unused$ was occupied or not. In the case of the latter, all subsequent missions utilizing said vehicle needed to be updated. These changes only held if they lowered the total Haversine distance and then the previously used (or occupied) base was transferred into $Unused$. Once all missions were explored, the iterators were reset and the algorithm repeated if there was further improvement found during the run.

The remaining changes occurred between lines lines 23-27 where vehicle at $Permutation_b$ index $i$ was compatible with the mission at $Permutation_b$ index $j$. The vehicles located at the respective indexed bases that would be assigned had to be of a compatible type (example: rotary-wing only as per Equation \ref{e10}) and were updated if they minimized the total Haversine distance. Additionally, if the replaced base was no longer assigned to any other mission, it was moved to $Unused$.

The parallelization of this algorithm was a bit more complex than the Base Ranking, as there were simultaneous accesses and dependencies. The outer loop at line 11 was eliminated and each index $i$ of the $Permutation_a$ matrix was assigned a thread. This would allow the CUDA platform to do simultaneous checks for each inner loop $j$ at once. In order to prevent race conditions a lock was placed in between lines 13 and 14, lines 18 and 19, and lines 23 and 24. Once the checks were completed, a thread was allowed to perform the adjustment. Additional threads would only act once the thread released the lock. This added a level of sequential accessing to the algorithm, although checks were performed in parallel and the lock would only activate upon a change occurring. Since updates did not occur as often as checks, much of the bottleneck of the algorithm was eliminated.

\begin{figure}[h!]
\centering
  \includegraphics[width=0.5\textwidth]{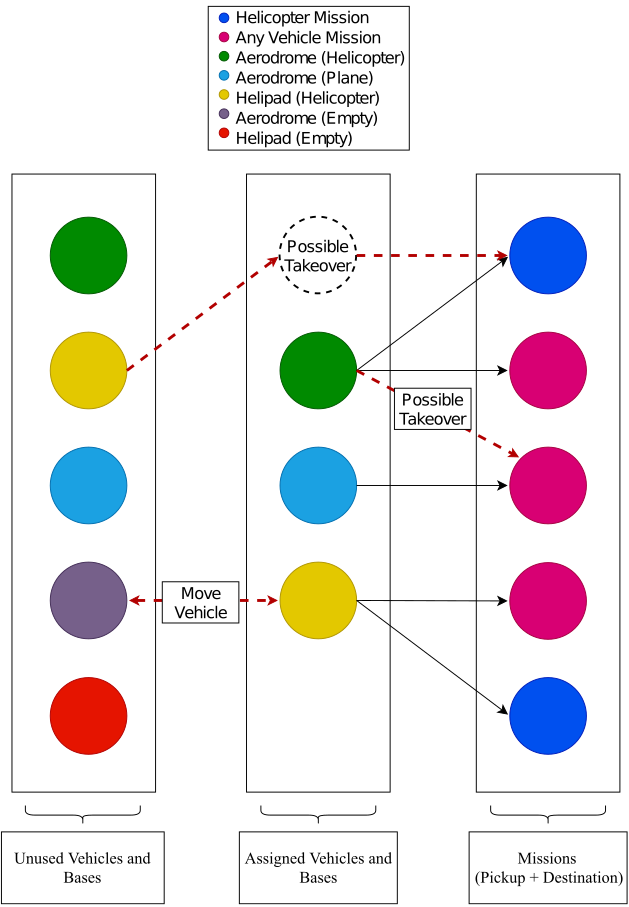}
\caption{Visualization of local search}
\label{fig_optimization}
\end{figure}

A visualization of this algorithm can be seen in Figure \ref{fig_optimization}. Assigned vehicles were applied to each mission (summed to pickup and destination), meeting Equations \ref{e3} to \ref{e10}. So long as the change was viable, another assigned base could attempt to take over the assignment of another. The new base would gain the mission, and the old base would be moved to $Unused$ if it had no more assignments remaining. A similar change took place based on whether an unused base contained a vehicle. If it did then a similar takeover occurred with the now assigned based moving from $Unused$. However, if the base did not contain a vehicle, then a swap occurred with the vehicle moving to the new location and the prior assigned base moving to $Unused$. Changes only held if they reduced the total Haversine distance applied across all missions.

\medskip
\begin{algorithm}[]
 \KwData{Ranked and Unused Mission Data From Algorithm \ref{base_rank}}
 \KwResult{Assignment of aircraft to missions and bases}
 $Destination \leftarrow$\ coordinates of destinations for respective missions\;
 $Pickup \leftarrow$\ coordinates of each mission\;
 $Base \leftarrow$\ coordinates of vehicle assigned to mission\;
 $Unused \leftarrow$\ coordinates of empty bases and unused vehicles\;
 $Permutation_a \leftarrow$\ random permutation of indices equal to the number of missions\;
 $Permutation_b \leftarrow$\ random permutation of indices equal to the number of missions\;
 $i \leftarrow$ first index of $Permutation_a$ vector\;
 $j \leftarrow$ first index of $Permutation_b$ vector\;
 $k \leftarrow$\ Total Haversine Distance\;
 \While{$improvement$}{
    \While{i not equal to the last index value in $Permutation_a$ vector}{
        \While{j not equal to the last index value in $Permutation_b$ vector}   {
           \If{unused vehicle $i$ is compatible with mission $j$}{
           Replace vehicle $j$ with unused vehicle $i$\;
           $l \leftarrow$\ Total Haversine Distance\;
           Update if $k$ is greater than $l$ and move vehicle $j$ to $Unused$ if not allocated to a mission;
           }
           \If{unused base $i$ can host vehicle $j$ and is compatible with mission $j$}{
           Move vehicle $j$ to unused base $i$\;
           $l \leftarrow$\ Total Haversine Distance\;
           Update if $k$ is greater than $l$ and move new empty based $j$ to $Unused$;
           }
           \If{vehicle $i$ is compatible with mission $j$}{
           Replace vehicle $j$ with vehicle $i$\;
           $l \leftarrow$\ Total Haversine Distance\;
           Update if $k$ is greater than $l$ and move vehicle $j$ to $Unused$ if not allocated to a mission;
           }
           $j =$ next index of $Permutation_b$ vector\;
        }
        $i =$ next index of $Permutation_a$ vector\;
    }
    $Permutation_a \leftarrow$\ new random permutation of indices equal to the number of missions\;
    $Permutation_b \leftarrow$\ new random permutation of indices equal to the number of missions\;
 }
 \caption{Local-Search Fleet Optimization}
 \label{local_search}
\end{algorithm}
\normalsize

\subsection{Tabu Search Fleet Optimization}
The Tabu search algorithm is an extension of Algorithm \ref{local_search}. For the most part, the operations have remained the same and even the parallelization operates in the same fashion. It still uses Algorithm \ref{base_rank} as a starting point. The primary contribution of this the introduction of the $TabuList$ at line 7. A Tabu search prevents recently explored neighbourhoods that improved the results from being explored again. The purpose of this is to give other possible changes a fair chance and prevent the algorithm from becoming stuck in a local optimum. Prior to each iteration $j$, a check is performed from lines 15-23. If the neighbourhood to be explored exists in the $TabuList$, then it is skipped depending on whether the selection is based on index $i$ or $j$ (lines 17-22). Selections are only added to the list if the improve the result as suggested by lines 28, 34, and 40. Each neighbourhood in the list has a counter, expressed by $TabuCounter$ and decremented at either line 16 or 42. After so many iterations, the selection is removed from the $TabuList$ and is allowed to be explored again.

It should be noted that a traditional Tabu search can potentially allow moves which will not improve the results. This was not done in the case of this algorithm, as it almost always resulted in a significantly poorer answer. One way to combat this problem was the introduction of more variability through the vectors $Permutation_a$ and $Permutation_b$. Originally both Algorithm \ref{local_search} and Algorithm \ref{tabu_search} did not utilize these which significantly impacted the performance. The two loops traversed sequentially through indices relative to the normal order the $Pickup$ matrix. As already suggested, this caused the results to be the exact same each time as Algorithm \ref{base_rank} would never have a different result. Additionally, using one permutation vector did improve the result, though the introduction of two allowed for significant variability and the exploration of previously unexplored neighbourhoods by older versions.

In terms of a parallel implementation, the algorithm performs almost identically to Algorithm \ref{local_search}. That being said, while the parallel local search results in the same answer as the sequential variation, the parallel Tabu search will not. The reasoning for this is due to the nature of Tabu search itself. In this algorithm neighbourhoods will only be explored if they are not in the list, otherwise they are skipped. Since multiple neighbourhoods are being explored simultaneously, the result of each can potentially be added to the list and the respective counters are reduced at different intervals. The forces it to explore differently that the non-parallel Tabu search, giving a comparable, yet different result. 

\medskip
\begin{algorithm}[]
 \KwData{Ranked and Unused Mission Data From Algorithm \ref{base_rank}}
 \KwResult{Assignment of aircraft to missions and bases}
 $Destination \leftarrow$\ coordinates of destinations for respective missions\;
 $Pickup \leftarrow$\ coordinates of each mission\;
 $Base \leftarrow$\ coordinates of vehicle assigned to mission\;
 $Unused \leftarrow$\ coordinates of empty bases and unused vehicles\;
 $Permutation_a \leftarrow$\ random permutation of indices equal to the number of missions\;
 $Permutation_b \leftarrow$\ random permutation of indices equal to the number of missions\;
 $TabuList \leftarrow$\ list of recently explored neighbourhoods\;
 $TabuCounter\leftarrow$\ how long a neighbourhood can be held within a list\;
 $i \leftarrow$ first index of $Permutation_a$ vector\;
 $j \leftarrow$ first index of $Permutation_b$ vector\;
 $k \leftarrow$\ Total Haversine Distance\;
 \While{$improvement$}{
    \While{i not equal to the last index value in $Permutation_a$ vector}{
        \While{j not equal to the last index value in $Permutation_b$ vector}   {
           \If{Selection $i$ or $j$ for $Unused$ or $Vehicle$ are in the $TabuList$}{
                Reduce the $TabuCounter$ for each within the $TabuList$\;
                \If{Selection is of an index $i$} {
                    Continue to next $i$ index of $Permutation_a$ vector\;
                }
                \Else{
                    Continue to next $j$ index of $Permutation_b$ vector\;
                }
           }
           \If{unused vehicle $i$ is compatible with mission $j$}{
           Replace vehicle $j$ with unused vehicle $i$\;
           $l \leftarrow$\ Total Haversine Distance\;
           Update if $k$ is greater than $l$ and move vehicle $j$ to $Unused$ if not allocated to a mission\;
           Add selection to the $TabuList$ and initiate $TabuCounter$ for the selection\;
           }
           \If{unused base $i$ can host vehicle $j$ and is compatible with mission $j$}{
           Move vehicle $j$ to unused base $i$\;
           $l \leftarrow$\ Total Haversine Distance\;
           Update if $k$ is greater than $l$ and move new empty based $j$ to $Unused$\;
           Add selection to the $TabuList$ and initiate $TabuCounter$ for the selection\;
           }
           \If{vehicle $i$ is compatible with mission $j$}{
           Replace vehicle $j$ with vehicle $i$\;
           $l \leftarrow$\ Total Haversine Distance\;
           Update if $k$ is greater than $l$ and move vehicle $j$ to $Unused$ if not allocated to a mission\;
           Add selection to the $TabuList$ and initiate $TabuCounter$ for the selection\;
           }
           Reduce the $TabuCounter$ for each within the $TabuList$\;
           $j =$ next index of $Permutation_b$ vector\;
        }
        $i =$ next index of $Permutation_a$ vector\;
    }
    $Permutation_a \leftarrow$\ new random permutation of indices equal to the number of missions\;
    $Permutation_b \leftarrow$\ new random permutation of indices equal to the number of missions\;
 }
 \caption{Tabu Search Fleet Optimization}
 \label{tabu_search}
\end{algorithm}
\normalsize

\section{RESULTS}
\label{sec:results}
Eleven datasets were analyzed for testing the previously described algorithms. All ran for 10 separate attempts; the results of which are summarized in Figures 2-5, Table 1, and Table 2. The datasets were generated as a randomized subset of known air and health facility coordinates. 12 vehicles (8 rotary-wing and 4 fixed-wing) and 378 bases (274 aerodromes and 104 helipads) were used for an individual assignment. Datasets initially ran through an instance of base ranking to generate a strong starting point, and then adjustments were made with the local search algorithm. The latter was performed until no further improvement could be found, at which point the results were recorded. As previously mentioned, this occurred ten times for each set with the average and limits being documented upon completion. Two separate instances were executed for each; one being sequential and the other being a parallel CUDA implementation. For validation, the Gurobi optimizer was employed to measure the algorithmic against the optimized solution. 

\begin{figure}[!h]
\centering
  \includegraphics[width=0.75\textwidth]{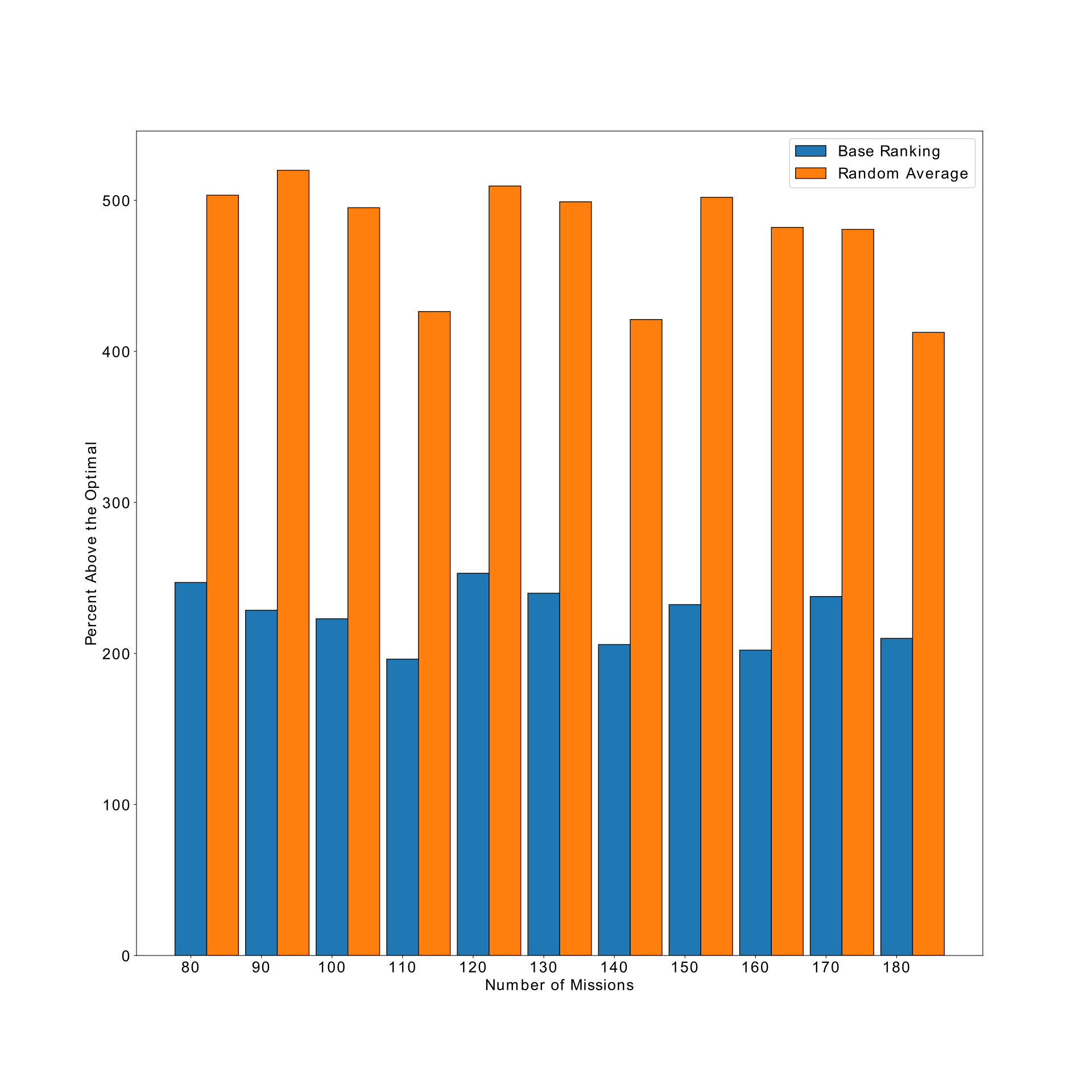}
\caption{Base Ranking versus random average starting over an optimal ground truth}
\label{fig_rank_local}
\end{figure}

\begin{figure}[!h]
\centering
  \includegraphics[width=0.75\textwidth]{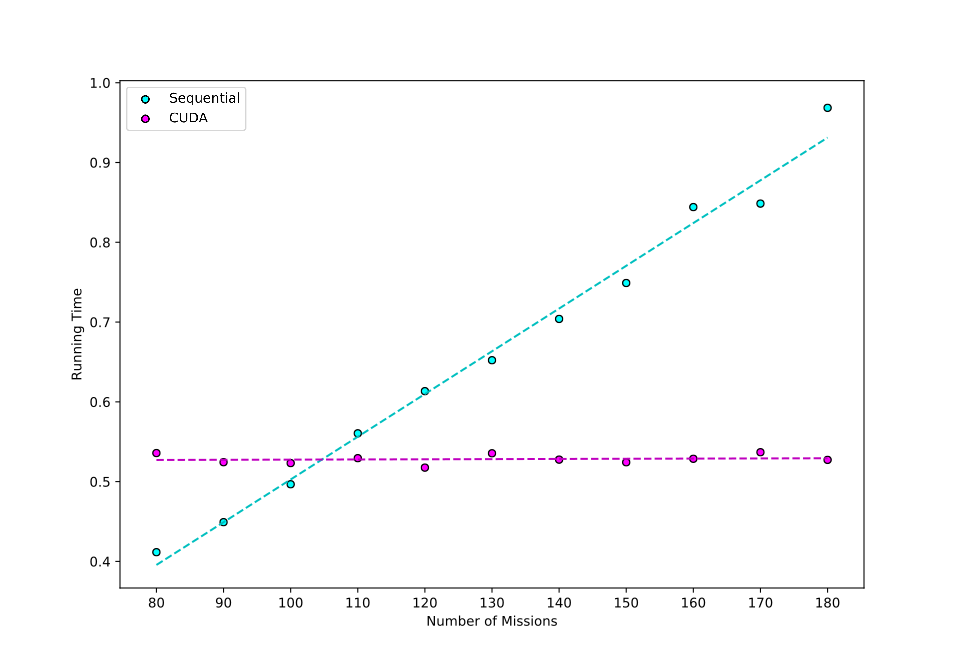}
\caption{Comparison of sequential and CUDA times for base ranking}
\label{fig_timing_rank}
\end{figure}

The purpose of the base ranking algorithm was to allow an improved starting position over a randomized permuted assignment. The results of this algorithm are displayed in Figure \ref{fig_rank_local} and Figure \ref{fig_timing_rank}. As there was no randomization in the operation, the runtimes and values were the same for every specific sequence. Figure \ref{fig_rank_local} compared this result against the average random starting position generated by 100 sets. In all cases the base ranking algorithm outperformed a random generation of bases applied to missions. While base ranking in conjunction with Tabu search allowed for a near-optimal result, it could not be used on its own for assignment. Despite being an improvement over randomization, Figure \ref{fig_rank_local} showed that base ranking was still substantially above the optimal ground truth. While the values were fairly uniform with an increase in mission size, they were not acceptable without additional algorithms. The speed of the base ranking should be noted, as it was far faster than the local or Tabu search. Furthermore, Figure \ref{fig_timing_rank} shows that the speed was nearly constant for the CUDA variation of the algorithm, while it increased linearly for the sequential version with the addition of missions. This result implies that for this scale the primary slow down point for the CUDA variation is the kernel call to the GPU itself. Given the consistency over randomization and the speed being negligible using CUDA meant it was worth performing upon the increasing sets. 


\begin{figure}[!h]
\centering
  \includegraphics[width=0.75\textwidth]{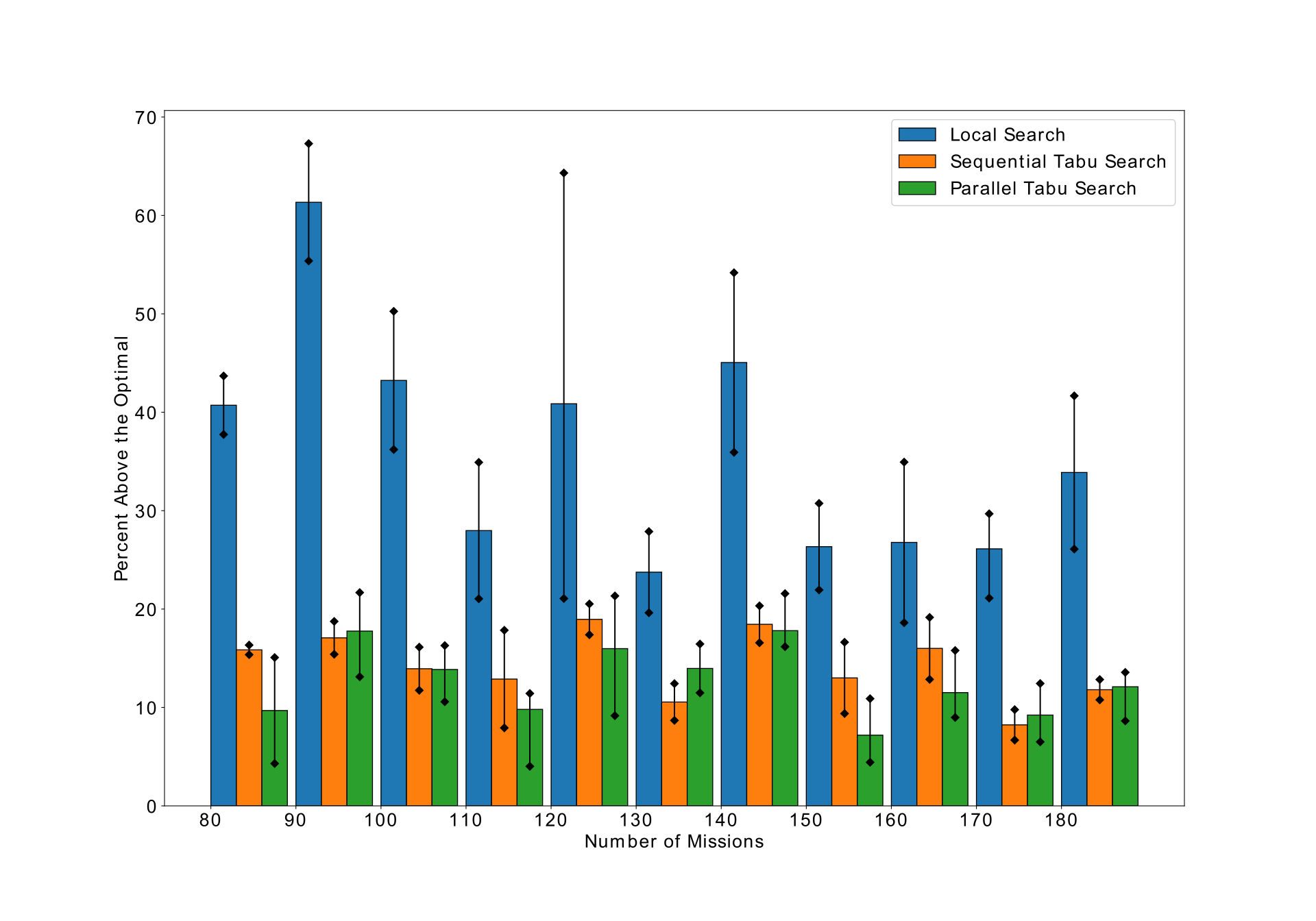}
\caption{Comparison of algorithms against the optimized solution}
\label{fig_results}
\end{figure}

On its own the local search algorithm proved unsuccessful in achieving a close to optimal solution. Figure \ref{fig_results} displays that even in the best scenarios, the increase was still just under 30\% and went as high as over 60\% for the average. The bounds were also not acceptable, with sets like 120 showing a very wide gap. These conclusions imply that it is getting stuck in a local optimum and the nature of the algorithm is preventing it from continuing. The Tabu search modification greatly improved the results and allowed it to achieve much closer to the Gurobi solutions. Per Figure \ref{fig_results}, all were consistently under 20\% and contained much smaller bounding gaps. This suggests that each solution was achieving a similar one to each other on successive tests. Likewise the parallel Tabu search was able to garner a similar result and even outperformed the sequential variation in some instances. There is no guarantee that the parallel version will always achieve a better metric to the sequential version, as they essentially use the same process with a difference mainly in runtime. However, they should always achieve a similar answer which is proven by Table \ref{fig_results_table}. It not only shows similar averages (A), but also comparable upper (U) and lower (L) bounds. It should be noted that there are still the possibility of outliers occurring, meaning that for real-use the algorithm should be performed multiple times.

\begin{figure}[!h]
\centering
  \includegraphics[width=0.75\textwidth]{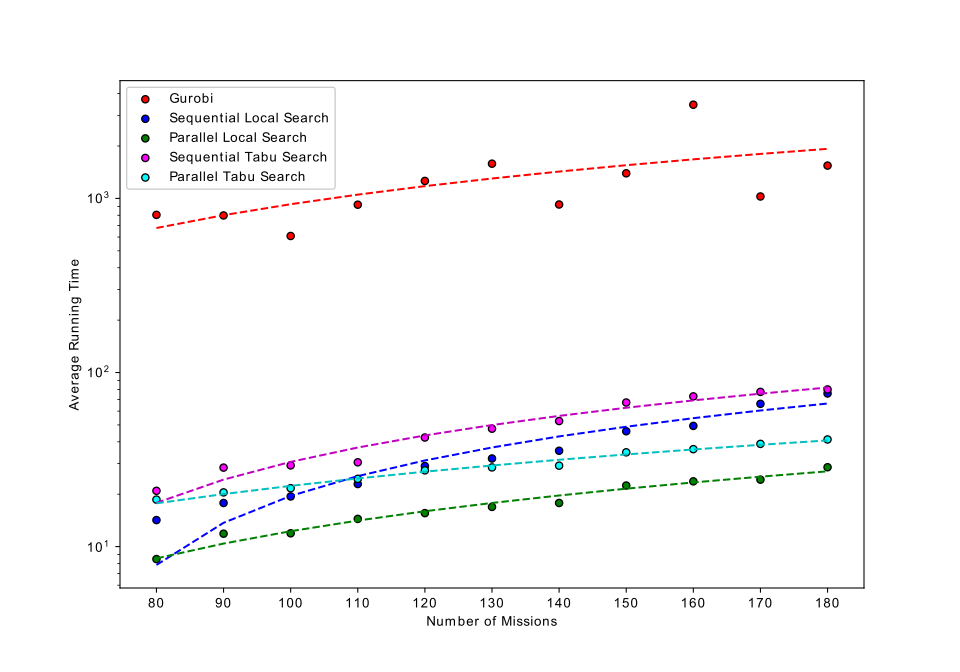}
\caption{Comparison of algorithmic runtimes (logarithmic scaling)}
\label{fig_results_time}
\end{figure}

\begin{table}[!h]
\centering
\caption{Summary of Total Haversine Distance Results.}
\label{fig_results_table}
\resizebox{0.75\textwidth}{!}{%
\begin{tabular}{|l|l|l|l|l|l|l|}
\hline
\textbf{Number of Missions} & \textbf{Optimal Distance} & \textbf{Local Search}                                                                 & \textbf{Tabu Search}                                                                  & \textbf{Parallel Tabu Search}                                                         \\ \hline
80                          & 19,665.9                 & \begin{tabular}[c]{@{}l@{}}U: 28,257.612\\ L: 27,087.806\\ A: 27,672.709\end{tabular} & \begin{tabular}[c]{@{}l@{}}U: 22,879.656\\ L: 22,687.638\\ A: 22,783.647\end{tabular} & \begin{tabular}[c]{@{}l@{}}U: 22,630.631\\ L: 20,512.258\\ A: 21,571.445\end{tabular} \\ \hline
90                          & 21,535.7                 & \begin{tabular}[c]{@{}l@{}}U: 36,028.446\\ L: 33,461.815\\ A: 34,745.130\end{tabular} & \begin{tabular}[c]{@{}l@{}}U: 25,572.773\\ L: 24,853.126\\ A: 25,212.949\end{tabular} & \begin{tabular}[c]{@{}l@{}}U: 26,203.804\\ L: 24,360.679\\ A: 25,372.478\end{tabular} \\ \hline
100                         & 23,578.5                 & \begin{tabular}[c]{@{}l@{}}U: 35,428.436\\ L: 32,116.723\\ A: 33,772.579\end{tabular} & \begin{tabular}[c]{@{}l@{}}U: 27,382.108\\ L: 26,346.672\\ A: 26,864.390\end{tabular} & \begin{tabular}[c]{@{}l@{}}U: 27,420.052\\ L: 26,073.513\\ A: 26,846.783\end{tabular} \\ \hline
110                         & 28,774.4                 & \begin{tabular}[c]{@{}l@{}}U: 38,819.082\\ L: 34,829.650\\ A: 36,824.366\end{tabular} & \begin{tabular}[c]{@{}l@{}}U: 33,909.447\\ L: 31,054.683\\ A: 32,482.065\end{tabular} & \begin{tabular}[c]{@{}l@{}}U: 32,061.554\\ L: 29,930.193\\ A: 31,595.873\end{tabular} \\ \hline
120                         & 27,338.4                 & \begin{tabular}[c]{@{}l@{}}U: 44,920.914\\ L: 33,098.822\\ A: 38,509.868\end{tabular} & \begin{tabular}[c]{@{}l@{}}U: 32,948.576\\ L: 32,092.066\\ A: 32,520.321\end{tabular} & \begin{tabular}[c]{@{}l@{}}U: 33,171.001\\ L: 29,843.666\\ A: 31,707.334\end{tabular} \\ \hline
130                         & 30,931.7                 & \begin{tabular}[c]{@{}l@{}}U: 39,554.642\\ L: 37,001.330\\ A: 38,277.991\end{tabular} & \begin{tabular}[c]{@{}l@{}}U: 34,774.973\\ L: 33,617.954\\ A: 34,196.463\end{tabular} & \begin{tabular}[c]{@{}l@{}}U: 36,019.660\\ L: 34,483.924\\ A: 35,251.797\end{tabular} \\ \hline
140                         & 37,530.2                 & \begin{tabular}[c]{@{}l@{}}U: 57,863.797\\ L: 51,014.197\\ A: 54,438.997\end{tabular} & \begin{tabular}[c]{@{}l@{}}U: 45,159.912\\ L: 43,750.540\\ A: 44,455.226\end{tabular} & \begin{tabular}[c]{@{}l@{}}U: 45,625.989\\ L: 43,599.954\\ A: 44,212.971\end{tabular} \\ \hline
150                         & 36,359.9                 & \begin{tabular}[c]{@{}l@{}}U: 47,536.625\\ L: 44,338.199\\ A: 45,937.412\end{tabular} & \begin{tabular}[c]{@{}l@{}}U: 42,406.212\\ L: 39,771.094\\ A: 41,088.653\end{tabular} & \begin{tabular}[c]{@{}l@{}}U: 40,321.420\\ L: 37,970.291\\ A: 38,972.549\end{tabular} \\ \hline
160                         & 40,578.4                 & \begin{tabular}[c]{@{}l@{}}U: 54,753.877\\ L: 48,131.950\\ A: 51,442.913\end{tabular} & \begin{tabular}[c]{@{}l@{}}U: 48,351.222\\ L: 45,793.889\\ A: 47,072.556\end{tabular} & \begin{tabular}[c]{@{}l@{}}U: 46,986.095\\ L: 44,223.194\\ A: 45,248.996\end{tabular} \\ \hline
170                         & 41,700.4                 & \begin{tabular}[c]{@{}l@{}}U: 54,078.253\\ L: 50,506.153\\ A: 52,592.203\end{tabular} & \begin{tabular}[c]{@{}l@{}}U: 45,778.813\\ L: 44,871.607\\ A: 45,133.210\end{tabular} & \begin{tabular}[c]{@{}l@{}}U: 46,886.254\\ L: 44,412.027\\ A: 45,549.141\end{tabular} \\ \hline
180                         & 49,758.6                 & \begin{tabular}[c]{@{}l@{}}U: 70,492.104\\ L: 62,744.421\\ A: 66,618.263\end{tabular} & \begin{tabular}[c]{@{}l@{}}U: 56,146.200\\ L: 55,119.247\\ A: 55,632.724\end{tabular} & \begin{tabular}[c]{@{}l@{}}U: 56,508.848\\ L: 54,055.691\\ A: 55,782.260\end{tabular} \\ \hline
\end{tabular}}
\end{table}
For the local search variation, the CUDA and sequential algorithms will always result in the same answer so long as matching permutations are used. As such, the only metric available for differentiating was time whic is summarized in Table \ref{fig_results_table_2}. The CUDA running time was a significant improvement over the sequential algorithms, only increasing by a small amount given the number of missions. In Figure \ref{fig_results_time} it can be seen that the trend line is much flatter than the sequential at a higher number of missions. This similarity is seen in the parallel Tabu search as well. Admittedly, it does display a comparable timing to the sequential variation at a lower mission count, though this changes with greater missions. It even manages to surpass the sequential local search after 110 missions. In all cases, the time for the sequential versions increased faster than the CUDA variations given an increase in the number. As previously speculated, this trend will continue to grow as the number of missions increases, giving validation to the CUDA implementation. Regardless, the timing in conjunction with near-optimal results justifies the use of parallelization for this purpose.

\begin{table}[!h]
\centering
\caption{Summary of Runtime Results (Seconds).}
\label{fig_results_table_2}
\resizebox{0.75\textwidth}{!}{%
\begin{tabular}{|l|l|l|l|l|l|}
\hline
\textbf{Missions} & \textbf{Gurobi (s)} & \textbf{Local Search (s)} & \textbf{Parallel Local Search (s)} & \textbf{Tabu Search (s)} & \textbf{Parallel Tabu Search (s)} \\ \hline
80                          & 804.503                 & 14.203                        & 8.478                                  & 20.915                       & 18.620                                \\ \hline
90                          & 798.339                 & 17.804                        & 11.863                                 & 28.426                       & 20.484                                \\ \hline
100                         & 608.483                 & 19.411                        & 11.925                                 & 29.272                       & 21.641                                \\ \hline
110                         & 920.3110                & 22.879                        & 14.424                                 & 30.491                       & 24.502                                \\ \hline
120                         & 1,260.200               & 29.026                        & 15.572                                 & 42.328                       & 27.415                                \\ \hline
130                         & 1,583.530               & 32.071                        & 16.909                                 & 47.558                       & 28.503                                \\ \hline
140                         & 922.661                 & 35.524                        & 17.800                                 & 52.625                       & 29.169                                \\ \hline
150                         & 1,394.010               & 45.977                        & 22.426                                 & 67.249                       & 34.803                                \\ \hline
160                         & 3,454.810               & 49.351                        & 23.705                                 & 72.937                       & 36.289                                \\ \hline
170                         & 1,025.990               & 66.105                        & 24.243                                 & 77.386                       & 38.909                                \\ \hline
180                         & 1,543.920               & 75.794                        & 28.569                                 & 79.921                       & 41.246                                \\ \hline
\end{tabular}}
\end{table}

\section{Conclusion}
\label{sec:conclusion}
The results generated demonstrates the use  fulness of the ranking and algorithmic solutions in both sequential and parallel forms. For the empirical data collected, Gurobi achieved an optimized solution in a significantly increased time. Though this time may be acceptable in cases where missions are known, it may not be viable in an emergency where reorganization is required. As such, a solution that delivers near optimization becomes all the more critical. All of the datasets reached a timeframe far exceeding traditional methodologies, while still being within an admissible target range. All algorithms approached an acceptable limit relative to the optimal, which was further enhanced, utilizing parallelization through the CUDA platform. On its own the local search proved insufficient, although modifying the algorithm into a Tabu search greatly enhanced the result. It should be noted that this model is adaptable to possible future changes in the data and could be updated quickly. This further denotes the advantage of these techniques over other similar solutions.


\bibliographystyle{unsrt}  
\bibliography{references}

\end{document}